\documentclass[conference]{IEEEtran}
\IEEEoverridecommandlockouts
\usepackage{cite}
\usepackage{amsmath,amssymb,amsfonts}
\usepackage{algorithmic}
\usepackage{graphicx}
\usepackage{lipsum}  
\usepackage{amsthm}
\usepackage{enumitem,kantlipsum}
\usepackage{url}
\DeclareMathOperator*{\argmin}{arg\,min}  

\newtheorem{theorem}{Theorem}

\usepackage{textcomp}
\usepackage{xcolor}
\def\BibTeX{{\rm B\kern-.05em{\sc i\kern-.025em b}\kern-.08em
    T\kern-.1667em\lower.7ex\hbox{E}\kern-.125emX}}
\begin{document}

\title{Revisiting Min-Max Optimization Problem in Adversarial Training\\

\thanks{Identify applicable funding agency here. If none, delete this.}
}

\author{\IEEEauthorblockN{Sina Hajer Ahmadi, Hassan Bahrami}
\IEEEauthorblockA{\textit{Electrical and Computer Engineering} \\
\textit{University of Waterloo}\\
Waterloo, Ontario}
}

\maketitle

\begin{abstract}
The rise of computer vision applications in the real world puts the security of the deep neural networks at risk. Recent works demonstrate that convolutional neural networks are susceptible to adversarial examples - where the input images look similar to the natural images but are classified incorrectly by the model. To provide a rebuttal to this problem, we propose a new method to build robust deep neural networks against adversarial attacks by reformulating the saddle point optimization problem in \cite{madry2017towards}. Our proposed method offers significant resistance and a concrete security guarantee against multiple adversaries. The goal of this paper is to act as a stepping stone for a new variation of deep learning models which would lead towards fully robust deep learning models.
\end{abstract}


\section{Introduction}
Convolutional Neural Networks (CNNs) are recently gaining popularity in numerous \cite{RealWorldApplicationsCNNS} computer vision tasks. Some examples include face detection, self-driving or autonomous cars and malware detection. The wide use of CNNs makes it prone to adversarial attacks and developing security measures against becomes crucial. These attacks are imperceptible to the naked eye and subtly attack the image in a way that only changes the image slightly but which is enough to fool the model completely. To design adversarial attack resistant models is becoming  a crucial design goal. 
\subsection{Literature Review}
Making the machine learning model more robust boils down to security \cite{biggio2018wild} reasons to protect it against adversarial attacks. Adversarial robust models can be interpreted \cite{santurkar2019image,tsipras2018robustness} better and can be generalized better \cite{bochkovskiy2020yolov4, zhu2019freelb}. Adversarial training deals with training the model on adversarial examples \cite{goodfellow2014explaining,madry2017towards }. Currently, machine learning practitioners favor adversarial training because (a) achieve higher robustness, (b) can be scaled to the state-of-the-art deep networks without affecting the inference time \cite{cohen2019certified} and (c) can deal with different attacks. Adversarial training can be pictured as a robust optimization problem \cite{shaham2018understanding, madry2017towards} which takes the form of a non-convex non-concave min-max problem. However, computing the optimal adversarial examples is an NP-hard problem \cite{andriushchenko2020understanding,weng2018towards }. In \cite{madry2017towards} adversarial training relies on approximation methods to solve the inner maximization problem. The PGD (projected gradient descent) attack \cite{madry2017towards} is successfully used where multiple steps of projected gradient descent are performed. It is widely accepted that models trained using  PGD  attack \cite{madry2017towards} are robust since small adversarial trained networks can be formally verified \cite{carlini2017provably,tjeng2017evaluating,wong2020fast }, and larger models could not be broken on public challenges \cite{madry2017towards,zhang2019theoretically,10446776}. \cite{croce2020reliable} evaluated the majority of recently published defenses and concluded that the standard $l_\infty$ PGD training achieves the best empirical robustness; a result which can only be improved using semi-supervised approaches \cite{hendrycks2019using,alayrac2019labels,carmon2019unlabeled}. In contrast to that  \cite{croce2020reliable} claim that over improving the standard PGD training had overestimated the robustness of their model. From all these experiments it is evident that adversarial training is the key algorithm for robust deep learning and performing it efficiently is of paramount importance. \cite{madry2017towards} observed that adversarial trained models do not perform well on adversarial perturbed test data. It was observed that there is a stark contrast and a huge gap between the training accuracy and the test accuracy on the adversarial data. From these differences and gaps it can be inferred that the training is severely over-fitting the model. 

In this paper, to tackle the above-mentioned problems, we reformulate the min-max optimization problem discussed in \cite{madry2017towards}. Unlike \cite{madry2017towards}, our method does not exploit the gradient of loss function for the inner maximization problem. Yet, it solves the inner maximization problem from the probabilistic perspective. To be more specific, first, a prior distribution for the allowed perturbation, namely $\delta$, is considered. Afterward, by sampling from this distribution, for each sample a sufficiently large number of attacked versions, say $k$, is generated. Moreover, the cross-entropy loss function, namely $L(.)$, is replaced by $e^{\lambda L(.)}$ for some large value of $\lambda$. As proven later, this will guarantee that the sum of loss function values for all of the $k$ attacked samples is proportional to the maximum loss value for that sample. This will resolve the problem of getting stuck in the local maximum obtained by PGD algorithm.  

We should also mention that the same idea has been deployed to train a fair federated learning model where the authors try to solve a multi objective optimization problem \cite{hu2020fedmgda,10381881,hamidi2024adafed}.

\subsection{Organization}
The organization of the paper is as follows. In section \ref{2}, we first briefly elaborate how \cite{madry2017towards} have performed adversarial training and then propose our own method that reformulates the saddle point optimization problem. In section \ref{3} we describe 3 different distributions to sample the perturbations from. Section \ref{4} summarizes the performance of our various models and compares them with the performance of the models in \cite{madry2017towards}. We conclude the paper in Section \ref{5} by making concrete comments over the results of our approach and talk briefly about our future work.

\section{Methodology \label{2}}
\subsection{MadryLab Adversarial Training}
In \cite{madry2017towards}, the authors formulated the adversarial training as the following saddle point optimization problem

\begin{equation}
\argmin_{\theta} \mathbb{E}_{(x,y)\sim \mathcal{D}} \{ \max_{\delta \in \mathcal{S}} L \left( \theta,x+\delta,y \right) \},
\end{equation}

where $\mathcal{S}$ is the allowed set of perturbation. Then, they performed various experiments focusing on two key elements: a) train a sufficiently high capacity network, b) use the strongest possible adversary in training.
In order to solve the inner maximization problem, they exploit projected gradient descent (PGD) starting from a random perturbation around the natural example. This algorithm effectively maximizes the loss of an example using only first order information. This approach demonstrated a steady decrease of training loss in the adversarial examples. They evaluated this trained model against a range of adversaries.
\subsection{Proposed Method}
In our proposed method, we reformulate the saddle point optimization problem in \cite{madry2017towards}.

\noindent\textbf{Assumptions:} we assume that loss function is smooth w.r.t. its inputs, and therefore w.r.t. $\delta$. 

\noindent\textbf{Notations:} $B_r(a)$ is an $n$-dimensional ball with radius $r$ centered at $a$. Denote by $\nabla L_x(.)$ and $\boldsymbol{H}_{x}$, the gradient vector and Hermitian matrix of loss function w.r.t. $x$, respectively.

\begin{theorem}\label{main_th}
For a large value of $\lambda$, the two following saddle point optimization problems are equivalent: \begin{equation}\label{MIT_eq}
\argmin_{\theta} \mathbb{E}_{(x,y)\sim \mathcal{D}} \{ \max_{\delta \in \mathcal{S}} L \left( \theta,x+\delta,y \right) \}
\end{equation}

\begin{equation}\label{our_eq}
\argmin_{\theta} \mathbb{E}_{(x,y)\sim \mathcal{D}} \{ \int_{\delta \in \mathcal{S}} e^{\lambda L \left( \theta,x+\delta,y \right)} d\delta \}
\end{equation}
\end{theorem}

\begin{proof}
To prove Theorem \ref{main_th}, it is sufficient to show that:
\begin{equation} \label{lem}
\max_{\delta \in \mathcal{S}} L \left( \theta,x+\delta,y \right)= \kappa \times \int_{\delta \in \mathcal{S}} e^{\lambda L \left( \theta,x+\delta,y \right)} d\delta,
\end{equation}
where $\kappa$ is a constant number. To this end, assume that $\delta^*$ yields the maximum value in \ref{lem}. Based on where the loss function attains its maximum value on $\mathcal{S}$, we consider the following cases (for notational simplicity, we denote $L \left( \theta,x+\delta,y \right)$ by $L \left(\delta\right)$):
\begin{enumerate}[wide, labelwidth=!, labelindent=0pt]

\item If there is a unique interior point $\delta^* \in \mathcal{S} $ attaining the maximum value. Since the loss function is smooth w.r.t. $\delta$, then $\nabla L \left(\delta^* \right)=0$. We prove that for large $\lambda$, 
\begin{align} \label{limit}
\frac{\int_{\delta \in \mathcal{S}} e^{\lambda L(\delta)} d \delta}{e^{\lambda L(\delta^*)} \sqrt{\frac{(2\pi)^n}{\lambda det(-\boldsymbol{H}_{\delta^*})}}}  \rightarrow 1,   
\end{align}
or equivalently, we prove that 
\begin{align} \label{limit2}
1 \leq \frac{\int_{\delta \in \mathcal{S}} e^{\lambda L(\delta)} d \delta}{e^{\lambda L(\delta^*)} \sqrt{\frac{(2\pi)^n}{\lambda det(-\boldsymbol{H}_{\delta^*})}}} \leq 1,
\end{align}
for some large $\lambda$. To this aim, we first establish upper and lower bounds for the loss function in the small vicinity around $\delta^*$. Based on the Taylor theorem, in the small ball $B_r(\delta^*)$ the loss function $L\left(\delta\right)$ is bounded by:

\begin{align} \label{taylor1}
L\left(\delta^*\right)+\frac{1}{2}(\delta - \delta^*)^T\boldsymbol{H}_{\delta^*}(\delta - \delta^*) - \mathcal{R}_2(\delta)< L\left(\delta\right) \nonumber \\
< L\left(\delta^*\right)+\frac{1}{2}(\delta - \delta^*)^T\boldsymbol{H}_{\delta^*}(\delta - \delta^*) + \mathcal{R}_2(\delta),
\end{align}
where $\mathcal{R}_2(\delta)$ is the second remainder of Taylor expansion. By defining
\begin{align}
\epsilon=\frac{\mathcal{R}_2(\delta)}{\max_{\delta \in B_r(\delta^*)} (\delta-\delta^*)^T(\delta-\delta^*)},
\end{align}
inequalities in \ref{taylor1} are simplified to
\begin{align} \label{taylor2}
L\left(\delta^*\right)+\frac{1}{2}(\delta - \delta^*)^T(\boldsymbol{H}_{\delta^*}-\epsilon)(\delta - \delta^*)< L\left(\delta\right) \nonumber \\
< L\left(\delta^*\right)+\frac{1}{2}(\delta - \delta^*)^T(\boldsymbol{H}_{\delta^*}+\epsilon)(\delta - \delta^*),
\end{align}
and therefore
\begin{align} \label{taylor3}
e^{\lambda\left(L\left(\delta^*\right)+\frac{1}{2}(\delta - \delta^*)^T(\boldsymbol{H}_{\delta^*}-\epsilon)(\delta - \delta^*)\right)}< e^{\lambda\left(L\left(\delta\right)\right)} \nonumber \\
< e^{\lambda\left(L\left(\delta^*\right)+\frac{1}{2}(\delta - \delta^*)^T(\boldsymbol{H}_{\delta^*}+\epsilon)(\delta - \delta^*)\right)}.
\end{align}
\noindent \textit{Lower Bound:} 
From the first inequality \ref{taylor3}
\begin{align}
&\int_{\delta \in \mathcal{S}} e^{\lambda\left(L\left(\delta\right)\right)} d \delta \nonumber \\ \label{eqq} &= \int_{\delta \in B_r(\delta^*)} e^{\lambda\left(L\left(\delta\right)\right)} d \delta + \int_{\delta \in \mathcal{S} \setminus B_r(\delta^*)} e^{\lambda\left(L\left(\delta\right)\right)} d \delta \\
& \geq \int_{\delta \in B_r(\delta^*)} e^{\lambda\left(L\left(\delta\right)\right)} d \delta \\ &\geq \int_{\delta \in B_r(\delta^*)} e^{\lambda\left(L\left(\delta^*\right)+\frac{1}{2}(\delta - \delta^*)^T(\boldsymbol{H}_{\delta^*}-\epsilon)(\delta - \delta^*)\right)} d \delta\\ 
&=e^{\lambda\left( L\left(\delta^*\right)\right)}\int_{\delta \in B_r(\delta^*)} e^{\frac{\lambda}{2} (\delta - \delta^*)^T(\boldsymbol{H}_{\delta^*}-\epsilon)(\delta - \delta^*)} d \delta \\ \label{yeq}
&=\frac{e^{\lambda\left( L\left(\delta^*\right)\right)}}{\sqrt{\lambda}} \int_{\delta \in B_{r\sqrt{\lambda}}(0)} e^{\frac{-1}{2} y^T(-\boldsymbol{H}_{\delta^*}+\epsilon)y} d y,
\end{align}
where \eqref{yeq} is obtained by setting $y=\sqrt(\lambda)(\delta - \delta^*)$. Note that since $L(\theta)$ is smooth and attained its maximum value in $\mathcal{S}$, $\boldsymbol{H}_{\delta^*}$ is negative definite in small vicinity around $\delta^*$, and therefore $-\boldsymbol{H}_{\delta^*}$ is positive definite. Thus,$-\boldsymbol{H}_{\delta^*}+\epsilon$ is also positive definite (maybe in smaller ball, I will make this more accurate). Hence, if $\lambda$ is large enough, by using multivariate normal integral, \eqref{yeq} is simplified to $\sqrt{\frac{(2\pi)^n}{det (-\boldsymbol{H}_{\delta^*}+\epsilon)}}$ (note that as $(-\boldsymbol{H}_{\delta^*}+\epsilon)$ is positive definite, its determinant is always positive). Hence,
\begin{align}
&\int_{\delta \in \mathcal{S}} e^{\lambda\left(L\left(\delta\right)\right)} d \delta \geq e^{\lambda L(\delta^*)} \sqrt{\frac{(2\pi)^n}{\lambda det(-\boldsymbol{H}_{\delta^*}+\epsilon)}}
\end{align}

\noindent \textit{Upper Bound:} 
Since $\delta^* \in B_r(\delta^*)$, for $\delta \in \mathcal{S} \setminus B_r(\delta^*)$, $L(\theta)<L(\delta^*)-\eta$ for some $\eta>0$. Therefore, 
\begin{align} \label{eqvol}
\int_{\delta \in \mathcal{S} \setminus B_r(\delta^*)} e^{\lambda\left(L\left(\delta\right)\right)} d \delta \leq \mathbf{Vol} \left(\mathcal{S} \setminus B_r  \left(\delta^*\right) \right)  \left(e^{\lambda\left( L(\delta^*)-\eta \right)}\right),
\end{align}
where $\mathbf{Vol}(.)$ denote the volume. Therefore, by using the second inequality in \eqref{taylor3}, and equations \eqref{eqq} and \eqref{eqvol}
\begin{align}
&\int_{\delta \in \mathcal{S}} e^{\lambda\left(L\left(\delta\right)\right)} d \delta \nonumber \\
& \nonumber \leq \int_{\delta \in B_r(\delta^*)} e^{\lambda\left(L\left(\delta^*\right)+\frac{1}{2}(\delta - \delta^*)^T(\boldsymbol{H}_{\delta^*}+\epsilon)(\delta - \delta^*)\right)} d \delta \\ &+ \mathbf{Vol} \left(\mathcal{S} \setminus B_r  \left(\delta^*\right) \right)  \left(e^{\lambda\left( L(\delta^*)-\eta \right)}\right)  \\ \nonumber
&=\frac{e^{\lambda\left( L\left(\delta^*\right)\right)}}{\sqrt{\lambda}} \int_{\delta \in B_{r\sqrt{\lambda}}(0)} e^{\frac{-1}{2} y^T(-\boldsymbol{H}_{\delta^*}-\epsilon)y} d y  \\  &+ \mathbf{Vol} \left(\mathcal{S} \setminus B_r  \left(\delta^*\right) \right)  \left(e^{\lambda\left( L(\delta^*)-\eta \right)}\right)
\\ &= e^{\lambda L(\delta^*)} \sqrt{\frac{(2\pi)^n}{\lambda \left( det(-\boldsymbol{H}_{\delta^*}-\epsilon ) \right)}}+ \mathbf{Vol} \left(\mathcal{S} \setminus B_r  \left(\delta^*\right) \right)  \left(e^{\lambda\left( L(\delta^*)-\eta \right)}\right).
\end{align}
If we divide both sides of the above inequality by 
\begin{align}
e^{\lambda L(\delta^*)} \sqrt{\frac{(2\pi)^n}{\lambda \left( det(-\boldsymbol{H}_{\delta^*}) \right)}},  
\end{align}
and then considering $\lambda$ is a large number, we obtain

\begin{align}
&\frac{\int_{\delta \in \mathcal{S}} e^{\lambda L(\delta)} d \delta}{e^{\lambda L(\delta^*)} \sqrt{\frac{(2\pi)^n}{\lambda det(-\boldsymbol{H}_{\delta^*})}}} \nonumber \\
& \leq \mathbf{Vol} \left(\mathcal{S} \setminus B_r  \left(\delta^*\right) \right) \left(e^{-\lambda \eta} \right) \sqrt{\frac{\lambda det(-\boldsymbol{H}_{\delta^*})}{(2\pi)^n}} \nonumber \\
&+\sqrt{\frac{det(-\boldsymbol{H}_{\delta^*})}{det(-\boldsymbol{H}_{\delta^*}-\epsilon)}}=\sqrt{\frac{det(-\boldsymbol{H}_{\delta^*})}{det(-\boldsymbol{H}_{\delta^*}-\epsilon)}},
\end{align}
which completes the proof for case 1.

\item If the maximum accrues at the edge of $\mathcal{S}$, then $\nabla L \left(\delta^* \right) \neq 0$. Then using the first order Taylor approximation we have
\begin{align}
L(\theta) \approx  L(\theta^*)+\nabla L(\theta^*)^T(\theta-\theta^*).    
\end{align}
Using the same method used for case 1, it could be proven that 
\begin{align} \label{limit2}
\frac{\int_{\delta \in \mathcal{S}} e^{\lambda L(\delta)} d \delta}{\frac{e^{\lambda L(\delta^*)}}{\lambda \left( \nabla L(\theta^*)\right)^n}}\rightarrow 1. 
\end{align}    

\item If there are $k$ absolute maximum occurring at multiple points, then \cite{10487854}
\begin{align}
\int_{\delta \in \mathcal{S}} e^{\lambda L(\delta)} d \delta=\left( \sum_{i=1}^k \kappa_i  \right) e^{\lambda L(\theta^*)}.  
\end{align}
\end{enumerate}
\end{proof}

\section{Sampling methods \label{3}}
To evaluate the integral in \eqref{our_eq}, we assume a prior distribution for $\delta \in \mathcal{S}$. In the following, we consider three different prior distributions.
\subsection{Uniform Sampling}
\begin{figure}[htbp]
\centerline{\includegraphics[width=7.43cm, height=5cm]{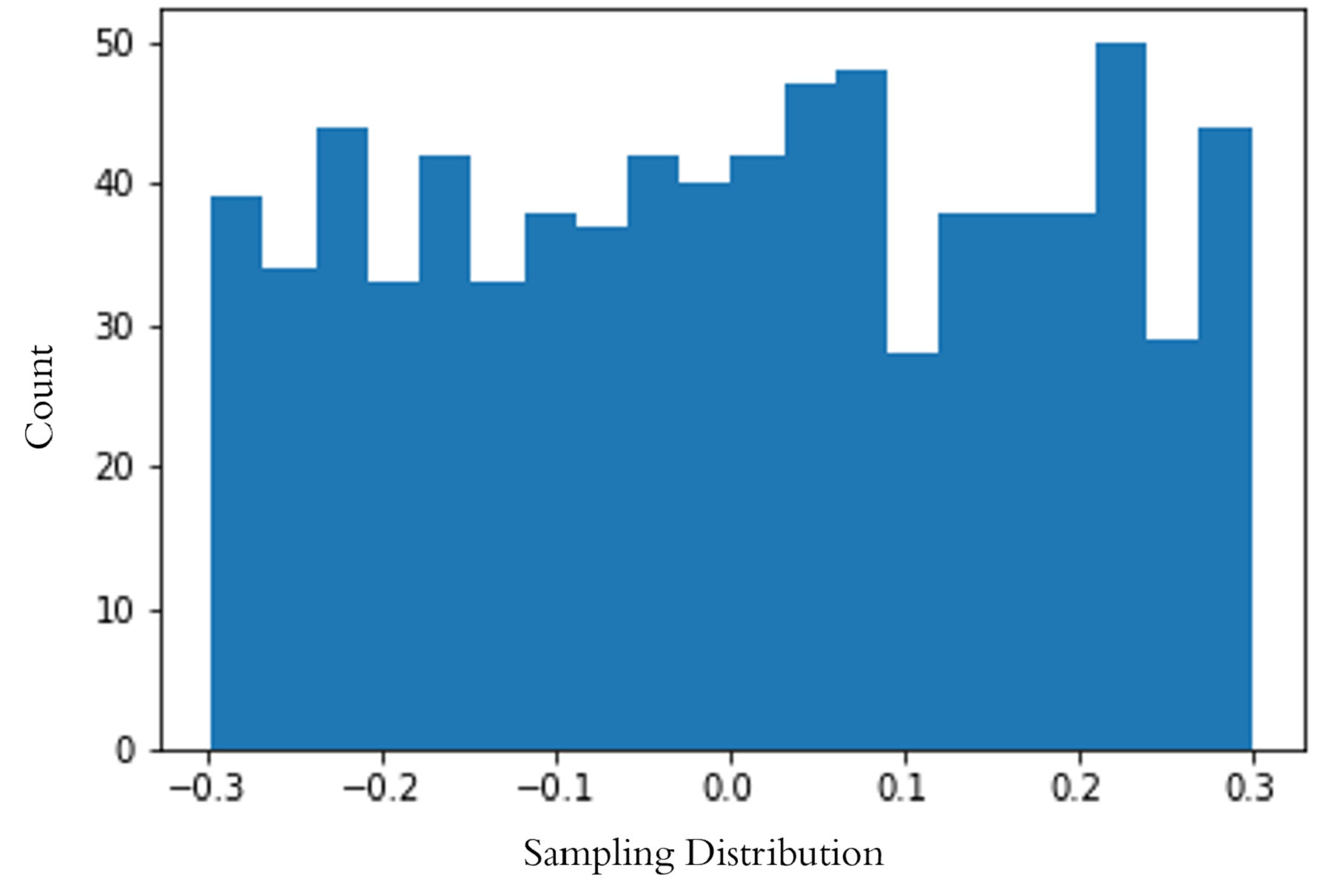}}
\caption{Uniform Distribution Sampling}
\label{fig}
\end{figure}
The first set of experiments that we carried out was sample the perturbations uniformly $-\epsilon$ to $+\epsilon$. We realised that the perturbations produced in this approach are fairly different than the perturbations produced when the network is attacked with different attacks like PGD and CW (Carlini and Wager).  We further tried to model the perturbations based on these attacks.

\subsection{Sampling Based on PGD and CW} \label{sub_sample}
In the next approach, we model the perturbations based on the PGD attack. To achieve this we used this formula:
\begin{equation}
Perturbations = X_{Attacked} - X_{orignal} 
\end{equation}
The distribution of the perturbations on a batch size of 50 images look like this:
\begin{figure}[htbp]
\centerline{\includegraphics[width=7.43cm, height=5cm]{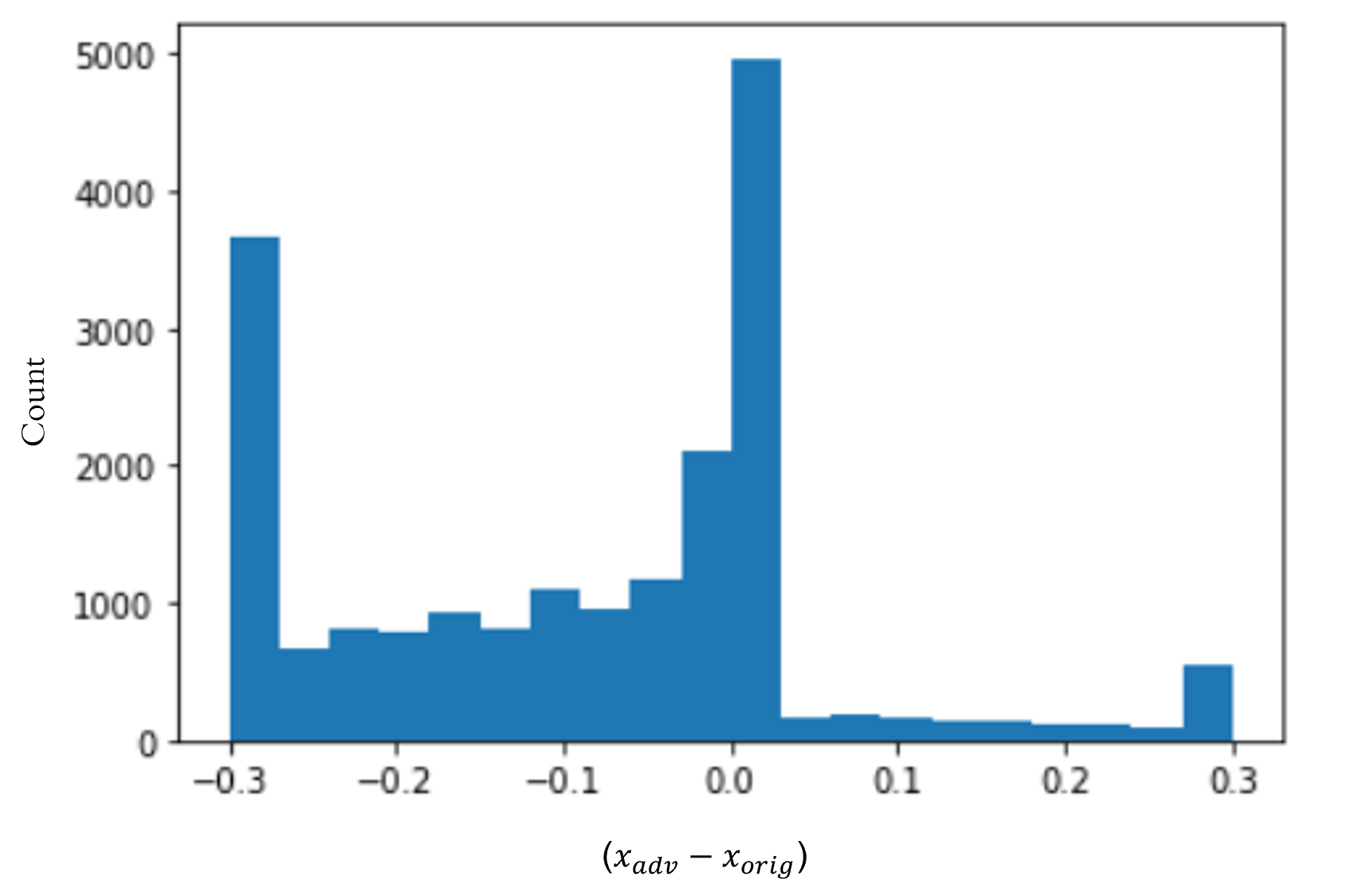}}
\caption{ Sampling based on PGD Perturbations}
\label{fig}
\end{figure}
Most of the values are concentrated between $-0.3$  and $0$ and the values have a spike for $-0.3$ and $0$. In our approach we sampled over these values and added these values to the image for training. 

We also modeled the distrubtion for the CW attack but ended up using the PGD attack distribution in our implementation.

\begin{figure}[htbp]
\centerline{\includegraphics[width=7.43cm, height=5cm]{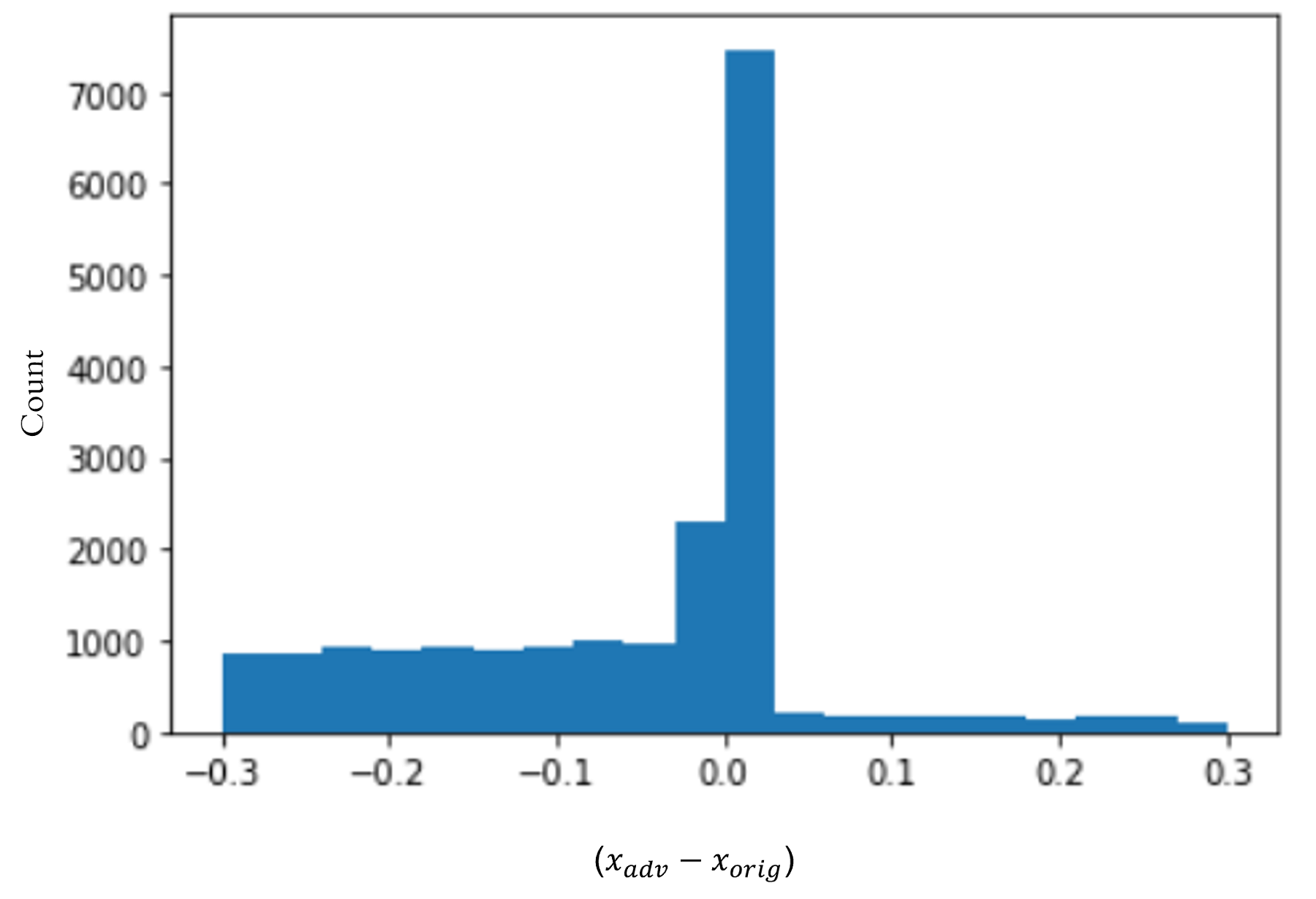}}
\caption{ Sampling based on CW Perturbations}
\label{fig}
\end{figure}
We observed that there is a uniform distribution from $-0.3$ to $0$ and most of the values are concentrated at $0$. 
\subsection{DCT Domain Sampling}
We also carried out two experiments in the DCT domain. The general idea of carrying out our experiments in the DCT domain is that instead of sampling the perturbation and adding them in the pixel domain, we sampled the perturbation and added them in the DCT domain. This was the process carried out by us:
\subsubsection{Sampling from Laplacian Distribution}
\begin{figure}[htbp]
\centerline{\includegraphics[width=8.55cm, height=2.5cm]{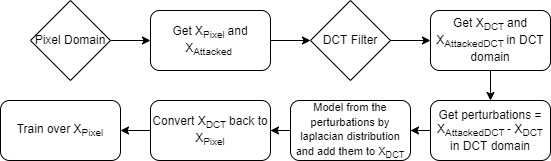}}
\caption{The diagram for simulating the perturbation in the DCT domain.}
\label{fig}
\end{figure}
\begin{itemize}
  \item We got $X_{Pixel}$ (Original Image) and $X_{Attacked}$ Images from the original Image.
  \item We passed them through a DCT filter and got their respectives in the DCT domain ($X_{DCT}$ and $X_{AttackedDCT}$).
  \item We then subtracted them to get the perturbations ($X_{AttackedDCT}$ - $X_{DCT}$) in the DCT domain.
  \item We then modeled these perturbations using the Laplacian distribution (see Fig. 5).
\begin{figure}[htbp]
\centerline{\includegraphics[width=7.5cm, height=5cm]{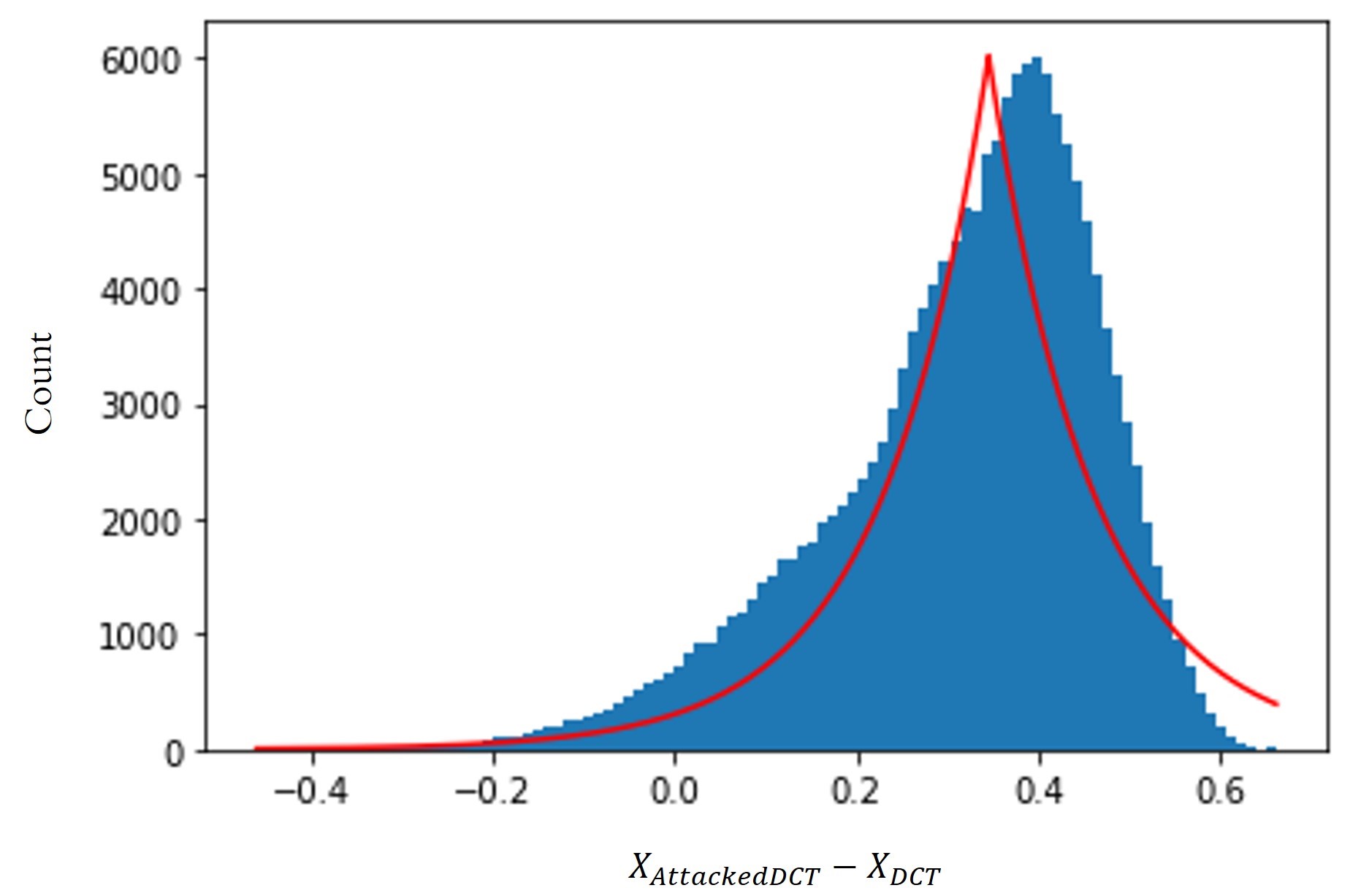}}
\caption{ The red line is the Laplacian distribution graph and the blue histogram is the empirical pdf of perturbation in DCT domain.}
\label{fig}
\end{figure}  
  \item We sampled from these perturbations and added them to $X_{DCT}$.
  \item Converted $X_{DCT}$ from the DCT domain back the $X_{Pixel}$ in the pixel domain.
  \item Trained the network over $X_{Pixel}$.
  
\end{itemize}

\subsubsection{Sampling from Empirical Probability Density Function}
The basic idea for this approach is very similar to the previous approach. In the previous approach we model the perturbations using the Laplacian distribution and then sample them these values in DCT domain. But in this approach we directly sample from the perturbations got in the DCT domain and add them to the image in the DCT domain. We then convert this image back to the pixel domain and train over these images. So the basic idea for this method is similar to the last one except the way of sampling is different.

\section{Experiments and Results \label{4}}
To justify the effectiveness of the proposed method, some experiments are carried out over the MNIST dataset. 

Based on the experiments we performed, the highest robustness achieved in our method is when we sample from the empirical pdf in either spacial or DCT domain. Hence, we only report the results for this type of sampling and compare the results with those reported in \cite{madry2017towards}. The number of samples used for our simulations is 100, so the batch size is 100 times that of the original one. In the following, first, we try different values for the exponential component $\lambda$ in the Spacial and the DCT domain for $\epsilon =\{0.1, 0.2, 0.3\}$, and compare the accuracy of our models with that of the \cite{madry2017towards}. Afterward, we compare the robustness of our model trained with images in the Spatial domain, with exponential component $\lambda$ = 1 and strength of the attack $\epsilon$ = 0.3 with the \cite{madry2017towards} model trained with $\epsilon$ = 0.3.

\subsection{Performance Analysis against PGD attack}
In this subsection we analyze the accuracy of our models trained a over $\epsilon$ = 0.1, 0.2 and 0.3.
\begin{enumerate}
\item $\epsilon = 0.1$

We analyze the performance of our model trained on $\epsilon$ = 0.1 and with exponential components $\lambda$ = 1,2 and compare it with the accuracy of the MadryLab model trained on $\epsilon$ = 0.1. See Figure 6 and Table 1.
\begin{figure}[htbp]
\centerline{\includegraphics[width=7.5cm, height=7.5cm]{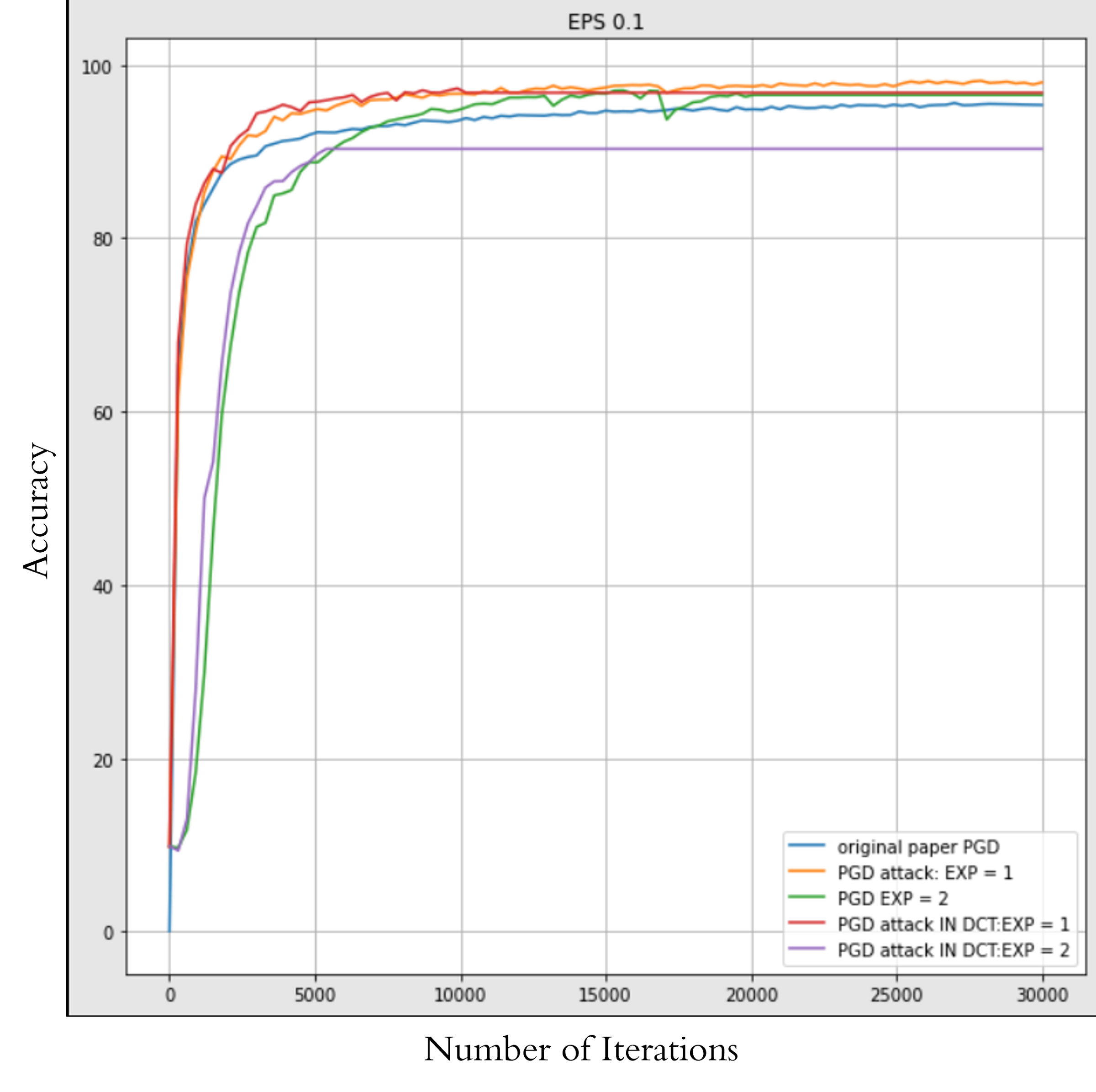}}
\caption{Performance of our adversarial trained models for exponential component $\lambda$ =1, 2 in the Spatial and DCT domain compared with the MadryLab adversarial trained model for $\epsilon$ = 0.1. }

\label{fig}
\end{figure}

\begin{table}[htbp]
\caption{Accuracy results for $\epsilon = 0.1$}
\begin{center}
\begin{tabular}{ |c|c|c|c| } 
 \hline
PGD &  $\lambda$  = 1 &  $\lambda$  = 2 & Madry Acc\\ 
  \hline
 MadryLab Model & - & - & 95.34\%\\ 
  \hline
 Spatial Domain &98.24\% & 98.24\% & - \\ 
 \hline
  DCT Domain &96.76\% & 90.28\% & - \\ 
 \hline
\end{tabular}
\label{tab1}
\end{center}
\end{table}
The observations noted for $\epsilon$ = 0.1 are that three of our trained models outperform the MadryLab trained model. For the exponential component $\lambda$ = 1, we outperform MadryLab with an accuracy of 98.24\% in the Spatial Domain and 96.76\% in the DCT Domain. For the exponential component $\lambda$ = 2, we outperfom the MadryLab only in the Spatial Domain with an accuracy of 96.5\%. The accuracy noted for the MadryLab model for $\epsilon$ = 0.1 is 95.34\%. 

The general trend of all the models seems similar and achieves a higher accuracy in the initial iterations and maintain that as the number of iterations increase.

\item $\epsilon = 0.2$

We analyze the performance of our models trained on $\epsilon$ = 0.2 and with exponential components $\lambda$ = 1,2,3,4 and compare it with the accuracy of the MadryLab model trained on $\epsilon$ = 0.2. See Figure 7 and Table 2.
\begin{figure}[htbp]
\centerline{\includegraphics[width=7.5cm, height=7.5cm]{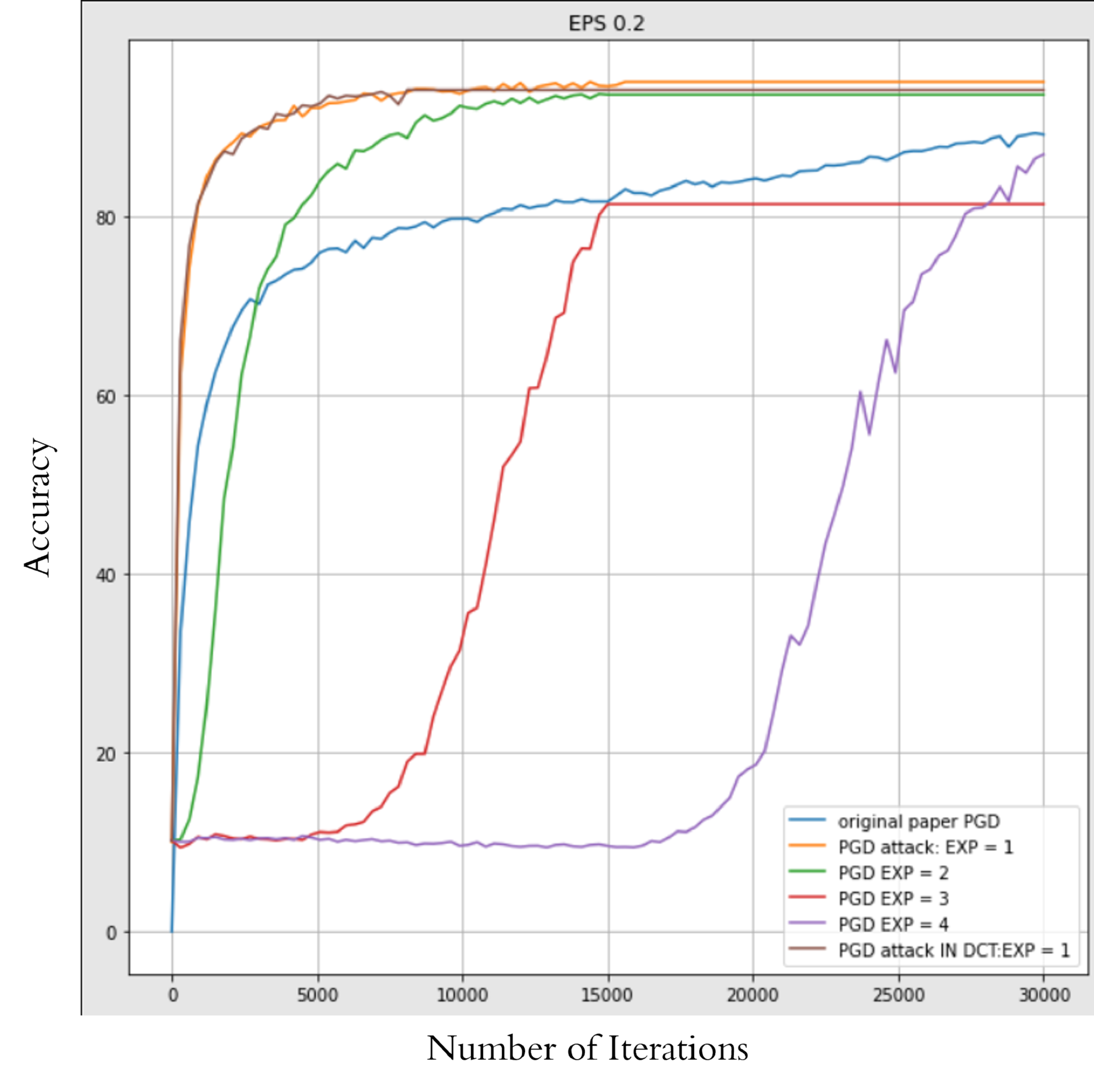}}
\caption{ Performance of our adversarial trained models for exponential component $\lambda$ =1, 2,3,4 in the Spatial and for exponential component $\lambda$ =1 in the DCT domain compared with the MadryLab adversarial trained model for $\epsilon$ = 0.2. }

\label{fig}
\end{figure}

\begin{table}[htbp]
\caption{Accuracy results for $\epsilon = 0.2$}
\begin{center}
\scalebox{0.9}{
\begin{tabular}{ |c|c|c|c|c|c| } 
 \hline
PGD &  $\lambda$  = 1 &  $\lambda$  = 2&  $\lambda$  = 3 &  $\lambda$  = 4& Madry Acc\\ 
  \hline
 MadryLab Model & - & - & - &- &89.17\%\\ 
  \hline
 Spatial Domain & 95.06\% & 93.64\% & 81.38 \% & 86.95\% &-\\ 
 \hline
  DCT Domain &94.18\% & - & -& -& - \\ 
 \hline
\end{tabular}}
\label{tab1}
\end{center}
\end{table}
For $\epsilon$ = 0.2, three of our trained models heavily outperform the Madrylab model. In the Spatial Domain, for the exponential component $\lambda$ = 1, 2, we achieved an accuracy of 95.06\% and 93.64\% respectively. In the DCT domain, for the exponential component $\lambda$ = 1, we achieved an accuracy of  94.18\%. Wheras the accuracy noted for the MadryLab model for $\epsilon$ = 0.2 is 89.17\%. 

The general trend of the curves is similar to the trend of the curves for $\epsilon$ = 0.1 except for when the  exponential component $\lambda$ was set at a higher value. It is noteworthy that when the exponential component $\lambda$ is higher ($\lambda$ = 2, 3) the accuracy is initially low but picks up exponentially. The accuracy curve for higher values of  $\lambda$ follows a similar trend for the exponential graph of its value.

\item $\epsilon = 0.3$

We analyze the performance of our models trained on $\epsilon$ = 0.3 and with exponential components $\lambda$ = 1,2,3,4 and compare it with the accuracy of the MadryLab model trained on $\epsilon$ = 0.3. See Figure 7 and Table 2.
\begin{figure}[htbp]
\centerline{\includegraphics[width=7.5cm, height=7.5cm]{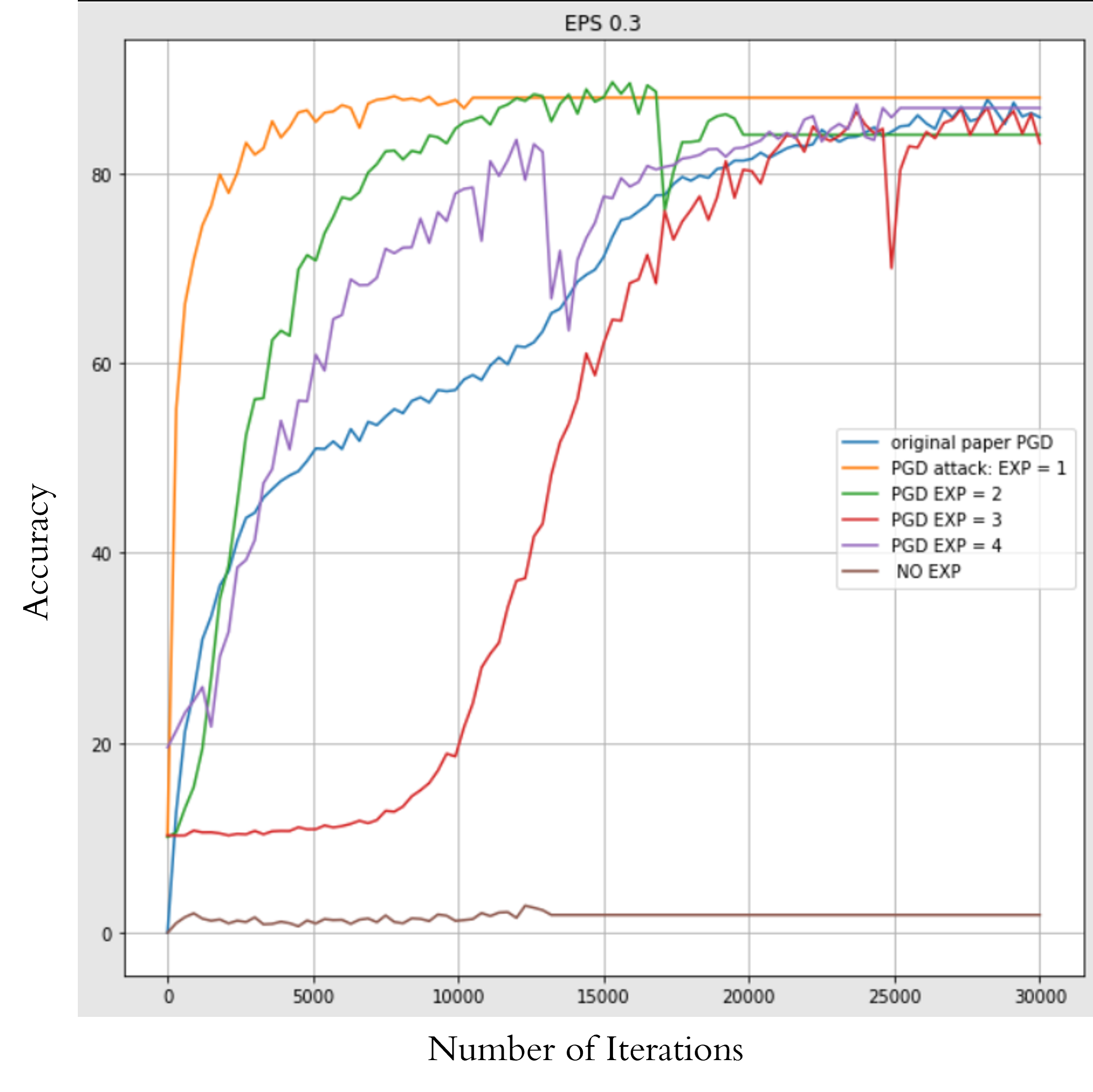}}
\caption{ Performance of our adversarial trained models for exponential component $\lambda$ =1, 2,3,4 in the Spatial compared with the MadryLab adversarial trained model for $\epsilon$ = 0.3. }

\label{fig}
\end{figure}

\begin{table}[htbp]
\caption{Accuracy results for $\epsilon = 0.3$}
\begin{center}
\scalebox{0.95}{

\begin{tabular}{ |c|c|c|c|c|c| } 
 \hline
PGD &  $\lambda$  = 1 &  $\lambda$  = 2&  $\lambda$  = 3 &  $\lambda$  = 4& Madry Acc\\ 
  \hline
 MadryLab Model & - & - & - &- &85.86\%\\ 
  \hline
 Spatial Domain & 87.92\% & 88.8\% & 83.89 \% & 87.22\% &-\\ 
 
 \hline
\end{tabular}}
\label{tab1}
\end{center}
\end{table}

For $\epsilon$ = 0.3, three of our trained models outperform the Madrylab model. In the Spatial Domain, for the exponential component $\lambda$ = 1, 2 and 4 we achieved an accuracy of 87.92\%, 88.8\% and 87.22\% respectively. Whereas the accuracy noted for the MadryLab model for $\epsilon$ = 0.3 is 85.86\%. 

The trend for $\epsilon$ = 0.3 for all models is quite erratic and for higher values of the model converges much slower and the graph is similar to the exponential graph of its  exponential component $\lambda$.

\end{enumerate}
\subsection{Robustness Check against PGD adversaries of different
strength}

\begin{figure}[htbp]
\centerline{\includegraphics[width=9cm, height=5cm]{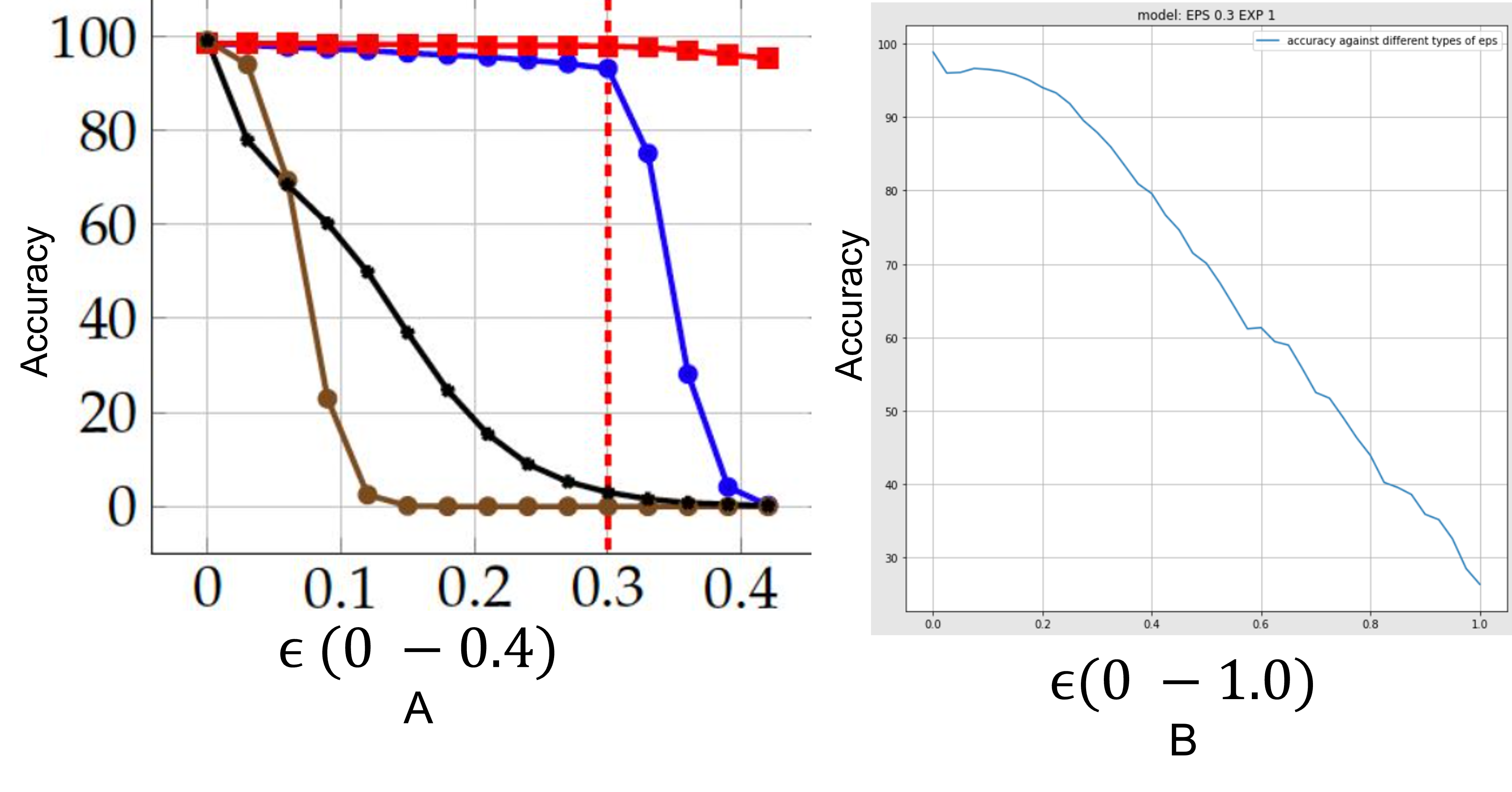}}
\caption{A: Performance of Madrylab network against PGD adversaries of different strength (blue line). B: Performance of our adversarial trained network against PGD adversaries of different strength. The model was trained against $\epsilon$ = 0.3. }

\label{fig}
\end{figure}
In this subsection we test the robustness of our model by testing it over a range of $\epsilon $ from 0 to 1. The model that we use is trained on $\epsilon$ = 0.3 and exponential component $\lambda$ = 1 and the training images are attacked in the Spatial Domain. We compare the robustness of our model with that of the MadryLab model.

In Figure 4a, for the Madrylab model, we observe that for $\epsilon$ less or equal to the value used during training, the performance is equal if not better. But when $\epsilon$ crosses its value for training, the accuracy plummets immediately and sinks to 0\%. Whereas, in Figure 4b, for our model, the accuracy gradually decreases for higher values of $\epsilon$. So, compared with the Madrylab model, our model is more robust and dependent against stronger attacks.  

\section{Conclusion \label{5}}
Our results suggest that deep neural network can be made resistant against adversarial attacks. Our theory of training the images over PGD sampled perturbations and augmenting the data and batch size is reliable. Making the loss exponential   aids our cause. We have proved that training the images by sampling them in both, spatial domain and DCT domain yields a high accuracy. Our different models both, in the spatial and DCT domain yield a higher accuracy that the model proposed by Madrylab. From the results, our adversarial models are more robust for stronger attacks than compared with the Madrylab model. 

Our model is very robust for the MNIST dataset and achieves a high accuracy against the PGD attack for a wide range of $\epsilon$ but does not do well against the CW attack. Our results show that, the methods that we have used lead to a significant increase in the robustness of the network. Our future work relies heavily on training and testing of the CIFAR10 dataset and testing our models on the FGSM attack.

\bibliographystyle{plain}
\bibliography{mybib}

\end{document}